\documentclass[10pt,conference]{IEEEtran}
\IEEEoverridecommandlockouts
\usepackage{cite}
\usepackage{amsmath,amssymb,amsfonts}
\usepackage{graphicx}
\usepackage{textcomp}
\usepackage{xcolor}

\usepackage{subcaption}
\usepackage{adjustbox}
\usepackage{makecell}
\usepackage{multirow}
\usepackage{bbding}
\usepackage{color, colortbl}
\usepackage{enumitem}

\usepackage{amsthm}

\usepackage[utf8]{inputenc} % allow utf-8 input
\usepackage[T1]{fontenc}    % use 8-bit T1 fonts
\usepackage{url}            % simple URL typesetting
\usepackage{booktabs}       % professional-quality tables
\usepackage{amsfonts}       % blackboard math symbols
\usepackage{nicefrac}       % compact symbols for 1/2, etc.
\usepackage{microtype}      % microtypography

\usepackage{circledsteps}

\usepackage{pifont}

\usepackage{algpseudocode}

\def\BibTeX{{\rm B\kern-.05em{\sc i\kern-.025em b}\kern-.08em
    T\kern-.1667em\lower.7ex\hbox{E}\kern-.125emX}}
\begin{document}

\title{Benchmarking Trustworthiness of SLMs: \\ Pre-trained vs. Compressed\\
}
% \begin{center}
\author{\IEEEauthorblockN{1\textsuperscript{st} Haokun Lin}
\IEEEauthorblockA{
\textit{Institute of Automation, CAS} \\
% \textit{City University of Hong Kong}\\
Beijing, China \\
ORCID 0009-0000-1084-7115}
\and
\IEEEauthorblockN{2\textsuperscript{nd} Kaijie Zhu}
\IEEEauthorblockA{\textit{Institute of Automation, CAS} \\
Beijing, China \\
ORCID 0009-0002-6220-1476}
\and
\IEEEauthorblockN{3\textsuperscript{rd} Haobo Xu}
\IEEEauthorblockA{\textit{Tsinghua University} \\
Beijing, China\\
ORCID 0009-0007-8311-7958}
\and
\IEEEauthorblockN{4\textsuperscript{th} Yichen Wu}
\IEEEauthorblockA{\textit{Harvard Medical School} \\
Boston, U.S. \\
ORCID 0000-0003-2859-3285}
\and

\IEEEauthorblockN{\hspace{0.55cm}5\textsuperscript{th} Zhichao Lu}
\IEEEauthorblockA{\hspace{0.55cm}
\textit{City University of Hong Kong}\\
\hspace{0.55cm}Hong Kong, China \\
\hspace{0.55cm}ORCID 0000-0002-4618-3573}
\and
\IEEEauthorblockN{6\textsuperscript{th} Qingfu Zhang}
\IEEEauthorblockA{\textit{City University of Hong Kong}\\
Hong Kong, China \\
ORCID 0000-0003-0786-0671}
\and
\IEEEauthorblockN{7\textsuperscript{th} Zhenan Sun}
\IEEEauthorblockA{\textit{Institute of Automation, CAS} \\
Beijing, China \\
ORCID 0000-0003-4029-9935}
}
% % \end{center}

\maketitle

\begin{abstract}

Small Language Models (SLMs) have emerged as a more efficient alternative to traditional Large Language Models (LLMs), offering promising potential in resource-constrained scenarios. Existing approaches to building SLMs typically follow two paths: training compact models from scratch, or compressing larger pre-trained models using methods such as pruning, quantization, or distillation. As language models become increasingly integrated into real-world applications, ensuring their trustworthiness has become a critical concern. However, how to build trustworthy SLMs remains an underexplored question. In this work, we present a comprehensive evaluation of SLM trustworthiness across multiple dimensions, including fairness, robustness, privacy, and ethics. We first examine the effects of pruning and quantization, and find that quantization is significantly more effective in preserving trustworthiness compared to pruning. More importantly, we demonstrate that compressing a reliable large model via quantization can produce SLMs with superior trustworthiness and adaptability compared to using small models trained from scratch. Furthermore, knowledge distillation from trustworthy teacher models can further enhance the reliability of SLMs. We hope our findings provide practical guidance and a foundation for future research into the development and deployment of trustworthy small language models.

\end{abstract}

\begin{IEEEkeywords}
Small language models, Pruning, Quantization, Knowledge distillation, Trustworthiness
\end{IEEEkeywords}

\section{Introduction}

Large Language Models (LLMs) have demonstrated remarkable performance across a wide range of natural language processing tasks~\cite{team2024gemma,qwen2.5,grattafiori2024llama}. This success is largely attributed to their massive parameter scales—models with over 7 billion parameters have become a popular baseline. However, deploying such large models incurs significant computational and memory costs, making them impractical for resource-constrained environments such as edge devices.
As a result, Small Language Models (SLMs), typically with fewer than 1–2 billion parameters, have attracted increasing attention for their efficiency during inference. There are two primary approaches to obtaining SLMs: (1) designing and training compact models from scratch using curated datasets and optimized architectures~\cite{thawakar2024mobillama,lozhkovsmollm2}, and (2) compressing larger LLMs via techniques like pruning, quantization, and distillation~\cite{frantar2022gptq,frantar2023sparsegpt,lin2023awq,lin2024duquant,lin2026efficient}. 
% While model compression provides a flexible and resource-efficient solution, performance degradation remains a major challenge.

\begin{figure*}[h]
    \centering
    % \hspace{-2em}
    \includegraphics[scale=0.6]{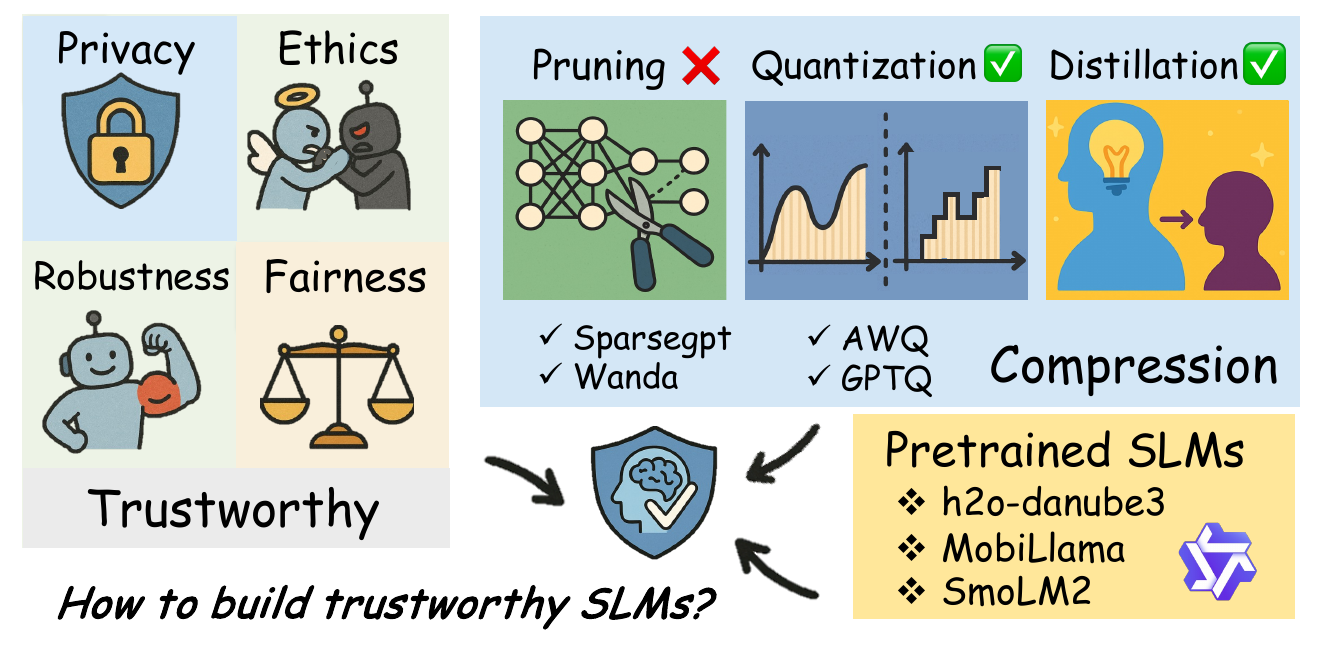}
    \caption{Our evaluation framework for assessing the trustworthiness of SLMs,
    including state-of-the-art pruning and quantization methods, a comparison between pre-trained SLMs and compressed larger models, and the impact of distillation.
    Our results suggest quantization as the preferred compression technique, while surpassing pre-trained SLMs.
    Distillation could provide additional improvements.
    }
    \label{fig:placeholder}
    \vspace{-0.3cm}
\end{figure*}

The growing deployment of LLMs in real-world applications brings trustworthiness to the forefront, involving aspects such as safety, fairness, robustness, privacy, and ethical alignment~\cite{chang2024survey}. 
Numerous studies~\cite{huang2024trustllm,zhu2023promptbench,das2025security} have investigated trustworthiness in LLMs, particularly for models exceeding 7B parameters or proprietary LLMs. 
However, how to ensure the trustworthiness of SLMs remains an open question. 
Prior works~\cite{egashira2024exploiting,hong2024decoding,chenassessing} have explored the effect of compression on trustworthiness, but few directly compare pre-trained SLMs with compressed larger models under a unified framework.
% leaving an important question largely unexplored.

In this work, we conduct a comprehensive analysis to understand how to build more trustworthy SLMs, using fairness, robustness, privacy, and ethics as key evaluation dimensions.
First, we evaluate the impact of pruning and quantization on model trustworthiness. Our findings across different model families and sizes suggest that pruning can impair reliability, whereas quantization largely preserves the trustworthiness of the original full-precision models. Therefore, we advocate using quantization as a more reliable path toward building SLMs.
Second, we directly compare several pre-trained SLMs (under 1B parameters) with quantized larger models. The results consistently show that quantized LLMs outperform pre-trained SLMs across all trustworthiness dimensions, indicating that compressing a reliable large model is more effective than training a small model from scratch.
Lastly, we explore the effect of knowledge distillation and find that distilling from a more trustworthy teacher can further enhance the reliability of SLMs.
Our main contributions are summarized as follows:
\begin{itemize}[leftmargin=5mm, itemsep=0mm, topsep=-0.5 mm]
    \item We conduct a thorough evaluation of compressed SLMs and recommend quantization over pruning as a more effective and reliable technique for preserving trustworthiness.
    \item We highlight a key insight: quantizing a larger, more trustworthy model yields more robust and flexible SLMs compared to directly using pre-trained small models. 
    % \item We discover that knowledge distillation can further enhance the trustworthiness of SLMs by transferring knowledge from stronger teacher models.
    \item We discover that knowledge distillation effectively improves SLM trustworthiness by leveraging the guidance of stronger teacher models.
\end{itemize}

\section{Related Work}

\subsection{Pre-Trained Small Language Models}
Recently, small language models (SLMs) have shown strong potential in resource-constrained scenarios compared to their large-scale counterparts.
Several works focus on designing SLMs from scratch. For example, MobiLlama~\cite{thawakar2024mobillama} improves efficiency by reducing redundancy in transformer blocks, while SmolLM2~\cite{lozhkovsmollm2} maximizes model performance through a multi-stage rebalancing of diverse training data sources.
In addition, the Qwen 2.5~\cite{qwen2.5} and Llama 3.2~\cite{grattafiori2024llama} series have released pre-trained SLMs with parameter scales ranging from 0.5B to 1B.
In this research, we explore the trustworthiness differences between such pre-trained SLMs and compressed LLMs.

\subsection{Model Compression}

\noindent
\textbf{Network Pruning} is a widely used compression technique that reduces model size by eliminating redundant or unimportant weights~\cite{han2015deep,xing2025efficientllm,xu2026prune}.
For large language models, unstructured pruning sets individual unimportant weights to zero, while N:M semi-structured pruning enforces a constraint that at least N out of every contiguous M weights must be zero. 
This semi-structured format is particularly advantageous on NVIDIA GPUs, as it enables acceleration of matrix multiply-accumulate operations.
Among representative methods, SparseGPT~\cite{frantar2023sparsegpt} improves the efficiency of the traditional Optimal Brain Surgeon (OBS) algorithm by adjusting the remaining weights to minimize the loss change caused by pruning. 
Wanda~\cite{sun2023wanda} proposes to leverage input activations as the importance indicator, while achieving performance comparable to SparseGPT.
Structured pruning~\cite{ma2023llmpruner} often causes notable performance degradation for LLMs without additional retraining; therefore, we do not include it in our exploration.

\noindent
\textbf{Network Quantization}
reduces the memory footprint of models by converting weight matrices into low-bit representations~\cite{yang2024dopq,lin2025quantization,yang2025lrq,zhang2026quantvla,lin2026duquant++,yang2026dapq,yang2026reshape}.
Post-training quantization has gained popularity for LLMs due to its low computational cost and the absence of retraining requirements.
GPTQ~\cite{frantar2022gptq} leverages second-order information to perform error compensation during quantization. 
AWQ~\cite{lin2023awq} identifies and protects salient weights with the assistance of the activation matrix, preserving model accuracy.
DuQuant~\cite{lin2024duquant} introduces rotation and permutation transformations to enhance low-bit weight-activation quantization.
% OmniQuant~\cite{shao2023omniquant} introduces lightweight trainable parameters to dynamically adjust the clipping ratio, improving quantization adaptability.

\noindent
\textbf{Knowledge Distillation} (KD) is a key technique for both compressing LLMs and improving their downstream performance.
The central idea is to transfer the knowledge encoded in a high-capacity teacher model into a smaller student model, typically by training the student to mimic the teacher’s output distributions~\cite{hinton2015distilling,xu2024slog,jiang2025image,xia2025medrek,li2026iv,yang2026mac}. 
% (e.g., logits or soft targets).
This paradigm enables compact models to retain much of the performance of their larger counterparts while significantly reducing computational overhead. 
Moreover, KD provides an effective means to extract task-specific or domain-specific knowledge from proprietary or closed-source models, serving as a practical alternative to direct access~\cite{xu2024survey_distill}.
In this work, we provide empirical insights that inform the trustworthy use of compressed models with these three techniques.
% we provide a unified study on how pruning, quantization, and distillation affect the robustness and safety of small LLMs.

\begin{table*}[!t]
    \centering
    \resizebox{0.6\linewidth}{!}{
\begin{tabular}{l|ccccc}
\toprule
\textbf{Model}                  & \textbf{Ethics} & \textbf{Privacy} & \textbf{Robustness} & \textbf{Fairness} & \textbf{Overall} \\ \midrule
\textbf{Gemma-1.1-7B}           & 82.19\%         & 60.13\%          & 68.90\%             & 44.73\%           & 63.99\%          \\ \cmidrule{2-6}
\textbf{Unstructured-Sparsegpt} & 81.12\%         & 58.40\%          & 63.84\%             & 38.43\%           & 60.45\%          \\
\textbf{Unstructured-Wanda}     & 80.90\%         & 56.86\%          & 64.41\%             & 29.82\%           & 58.00\%          \\ \cmidrule{2-6}
\textbf{4:8-Sparsegpt}          & 77.36\%         & 59.24\%          & 65.20\%             & 32.66\%           & 58.61\%          \\
\textbf{4:8-Wanda}              & 74.87\%         & 52.70\%          & 63.57\%             & 54.50\%           & 61.41\%          \\ \cmidrule{2-6}
\textbf{2:4-Sparsegpt}          & 72.92\%         & 56.37\%          & 61.82\%             & 24.43\%           & 53.89\%          \\
\textbf{2:4-wanda}              & 73.08\%         & 55.35\%          & 59.27\%             & 36.23\%           & 55.98\%          \\ \midrule
\textbf{Llama-3.1-8B}           & 81.29\%         & 49.16\%          & 71.62\%             & 36.03\%           & 59.52\%          \\ \cmidrule{2-6}
\textbf{Unstructure-Sparsegpt}  & 80.05\%         & 46.08\%          & 60.57\%             & 44.08\%           & 57.69\%          \\
\textbf{Unstructured-Wanda}     & 77.18\%         & 51.56\%          & 61.04\%             & 31.76\%           & 55.38\%          \\ \cmidrule{2-6}
\textbf{4:8-Sparsegpt}          & 76.74\%         & 53.40\%          & 54.87\%             & 44.10\%           & 57.28\%          \\
\textbf{4:8-Wanda}              & 74.98\%         & 48.67\%          & 57.29\%             & 46.19\%           & 56.78\%          \\ \cmidrule{2-6}
\textbf{2:4-Sparsegpt}          & 66.02\%         & 47.75\%          & 56.76\%             & 44.65\%           & 53.79\%          \\
\textbf{2:4-Wanda}              & 59.61\%         & 42.22\%          & 47.27\%             & 72.04\%           & 55.29\%          \\ \midrule
\textbf{Qwen2.5-7B}             & 83.78\%         & 41.53\%          & 67.05\%             & 28.33\%           & 55.17\%          \\ \cmidrule{2-6}
\textbf{Unstructured-Sparsegpt} & 84.29\%         & 44.22\%          & 58.64\%             & 29.51\%           & 54.17\%          \\ 
\textbf{Unstructured-Wanda}     & 84.42\%         & 38.38\%          & 61.01\%             & 28.79\%           & 53.15\%          \\\cmidrule{2-6}
\textbf{4:8-Sparsegpt}          & 83.99\%         & 40.87\%          & 60.27\%             & 30.42\%           & 53.89\%          \\ 
\textbf{4:8-Wanda}              & 83.86\%         & 38.41\%          & 56.35\%             & 36.14\%           & 53.69\%          \\ \cmidrule{2-6}
\textbf{2:4-Sparsegpt}          & 82.12\%         & 40.36\%          & 53.37\%             & 30.65\%           & 51.62\%          \\
\textbf{2:4-Wanda}              & 82.20\%         & 39.48\%          & 56.48\%             & 29.89\%           & 52.01\%          \\ \bottomrule
\end{tabular}
    }
    \caption{Trustworthiness assessment for pruned LLMs.}
    \label{tab:pruning}
    \vspace{-0.3cm}
\end{table*}

\subsection{Model Trustworthiness}

TrustLLM~\cite{huang2024trustllm} evaluates the reliability of language models across six key dimensions. Truthfulness measures whether a model conveys accurate information. Safety focuses on preventing harmful, unsafe, or unlawful outputs. Fairness ensures that models do not introduce bias or discrimination across different demographics. Robustness captures a model’s ability to maintain stable performance under diverse inputs or conditions. Privacy emphasizes protecting individual autonomy and sensitive data. Together, these dimensions provide a comprehensive framework for assessing the trustworthiness of LLMs.
For SLMs, prior research~\cite{egashira2024exploiting} has mainly examined the safety of quantized LLMs, while \cite{hong2024decoding} focuses on the trustworthiness of compressed large models.
In contrast, our work investigates the trustworthiness of SLMs by directly comparing pre-trained SLMs with compressed larger models.

\section{Preliminary}

\noindent
\textbf{Quantization.} 
The general $b$-bit uniform quantization $\mathcal{Q}_b(\cdot)$ can be represented as:
\begin{equation}
\label{eq:quant}
    \hat{\mathbf {x}} = \mathcal{Q}_b(\mathbf {x}) = s \cdot \Pi_{\Omega(b)}(\mathbf{x}/s),
\end{equation}
where $s$ is the quantization step size, and $\Pi_{\Omega(b)}$ is the projection function onto the set of $b$-bit integers $\Omega(b) = \{0, 1, ..., 2^{b}-1\}$.

\noindent
\textbf{Knowledge Distillation.}
% We typically 
Given an input $x$, the teacher model produces logits $z^{(T)}$ and the student model produces $z^{(S)}$. 
With temperature $T$, the softened probabilities are
\begin{equation}
p_i^{(T)} = \frac{\exp(z_i^{(T)}/T)}{\sum_j \exp(z_j^{(T)}/T)}, \quad 
q_i = \frac{\exp(z_i^{(S)}/T)}{\sum_j \exp(z_j^{(S)}/T)}.
\end{equation}
The supervised distillation~\cite{hinton2015distilling} minimizes the KL divergence:
\begin{equation}
\mathcal{L}_{\text{KD}} = \sum_i p_i^{(T)} \log \frac{p_i^{(T)}}{q_i}.
\end{equation}

\section{Safety Evaluation for Small LLMs}

In this section, we present a series of experiments to address our core research question: \textbf{How can we obtain more trustworthy small language models?} To systematically explore this, we further investigate the following sub-questions:

\begin{itemize}[leftmargin=5mm, itemsep=0mm, topsep=-0.5 mm]
    \item \textbf{RQ1}: Are pruned large language models trustworthy?
    \item \textbf{RQ2}: Are quantized large language models trustworthy?
    \item \textbf{RQ3}: Are compressed LLMs more trustworthy than pre-trained SLMs?
    \item \textbf{RQ4}: Can distillation improve trustworthiness?
\end{itemize}

\begin{table*}[!t]
    \centering
    \resizebox{0.6\linewidth}{!}{
\begin{tabular}{l|ccccc}
\toprule
\textbf{Model}         & \textbf{Ethics} & \textbf{Privacy} & \textbf{Robustness} & \textbf{Fairness} & \textbf{Overall} \\ \midrule
\textbf{Llama-3.2-1B}   & 73.98\%         & 60.21\%          & 59.43\%             & 70.81\%           & 66.11\%          \\
\textbf{1B-INT4-AWQ}    & 69.71\%         & 59.71\%          & 54.17\%             & 79.11\%           & 65.67\%          \\
\textbf{1B-INT4-GPTQ}   & 72.32\%         & 57.74\%          & 56.19\%             & 68.45\%           & 63.67\%          \\ \midrule
\textbf{Llama-3.2-3B}   & 82.88\%         & 60.15\%          & 66.42\%             & 43.30\%           & 63.19\%          \\
\textbf{3B-INT4-AWQ}    & 81.27\%         & 56.58\%          & 58.80\%             & 56.51\%           & 63.29\%          \\
\textbf{3B-INT4-GPTQ}   & 81.93\%         & 57.53\%          & 67.21\%             & 59.52\%           & 66.55\%          \\ \midrule
\midrule
\textbf{Qwen2.5-0.5B}   & 65.71\%         & 27.56\%          & 62.42\%             & 48.99\%           & 51.17\%          \\
\textbf{0.5B-INT4-AWQ}  & 57.56\%         & 27.77\%          & 53.34\%             & 48.67\%           & 46.83\%          \\
\textbf{0.5B-INT4-GPTQ} & 69.26\%         & 27.91\%          & 62.17\%             & 52.37\%           & 52.93\%          \\ \midrule
\textbf{Qwen2.5-1.5B}   & 77.52\%         & 40.89\%          & 69.38\%             & 71.82\%           & 64.90\%          \\
\textbf{1.5B-INT4-AWQ}  & 73.83\%         & 38.97\%          & 68.64\%             & 71.89\%           & 63.33\%          \\
\textbf{1.5B-INT4-GPTQ} & 78.63\%         & 40.38\%          & 67.34\%             & 69.07\%           & 63.86\%          \\ \midrule
\textbf{Qwen2.5-3B}     & 80.13\%         & 40.48\%          & 68.79\%             & 29.99\%           & 54.85\%          \\
\textbf{3B-INT4-AWQ}    & 79.00\%         & 40.93\%          & 68.49\%             & 32.10\%           & 55.13\%          \\
\textbf{3B-INT4-GPTQ}   & 80.43\%         & 44.98\%          & 68.75\%             & 29.82\%           & 56.00\%          \\ \midrule
\textbf{Qwen2.5-7B}     & 83.78\%         & 41.53\%          & 67.05\%             & 28.33\%           & 55.17\%          \\
\textbf{7B-INT4-AWQ}    & 83.79\%         & 39.02\%          & 67.68\%             & 28.71\%           & 54.80\%          \\
\textbf{7B-INT4-GPTQ}   & 83.87\%         & 40.15\%          & 68.09\%             & 29.18\%           & 55.32\%          \\ \bottomrule
\end{tabular}
    }
    \caption{Trustworthiness assessment for quantized LLMs.}
    \label{tab:quantization}
    % \vspace{-0.3cm}
\end{table*}

\subsection{Setup}

\noindent
\textbf{Language Models and Compression Methods.}
We conduct comprehensive evaluations across a range of instruction-tuned language models. 
For pre-trained small language models, we select the following models: h2o-danube3-500m-Chat~\cite{pfeiffer2024h2o}, MobiLlama-500m-Chat~\cite{thawakar2024mobillama}, SmolLM2-360M-Instruct~\cite{lozhkovsmollm2}, and Qwen2.5-0.5B-Instruct~\cite{qwen2.5}.
For compression methods, we adopt widely used techniques, including SparseGPT~\cite{frantar2023sparsegpt} and Wanda~\cite{sun2023wanda} for pruning, as well as GPTQ~\cite{frantar2022gptq} and AWQ~\cite{lin2023awq} for quantization. All compression experiments are calibrated using the WikiText-v2 dataset.
To obtain compressed SLMs, we apply these compression methods to commonly used LLMs, such as Gemma~\cite{team2024gemma}, Llama~\cite{grattafiori2024llama}, and Qwen~\cite{qwen2.5}.
For distillation, we conduct supervised knowledge distillation~\cite{hinton2015distilling} on Qwen2.5 models using Alpaca dataset~\cite{bommasani2021opportunities}.
% domain-specific datasets in code and math.

\noindent
\textbf{Trustworthiness Measurement.}
We adopt the TrustLLM~\cite{huang2024trustllm} benchmark as the primary framework for assessing trustworthiness. Specifically, we evaluate models across four dimensions: machine ethics, privacy, robustness, and fairness.
For ethics, we use the ETHICS~\cite{hendrycks2020ethics} and Social-Chem-101~\cite{forbes2020social} datasets to assess implicit ethics, and the MoralChoice dataset for explicit ethics evaluation.
For privacy, we adopt agreement tests on private information usage and privacy scenario tasks.
For robustness, we use AdvGLUE~\cite{wang2021advbench} and AdvInstruction to measure resistance to natural noise, and further examine performance on out-of-distribution (OOD) detection and generalization tasks.
For fairness, we evaluate from three perspectives: disparagement, representation bias, and preference bias in subjective choices.
We report the average accuracy on sub-tasks under each dimension, and the overall average score is used as the indicator of model trustworthiness.

\subsection{Pruning Influence}

\noindent
\textbf{Pruning generally harms trustworthiness.}
We conduct comprehensive evaluations of both unstructured and semi-structured pruning on various large language models, particularly those at the 7B scale. 
The results, summarized in Table~\ref{tab:pruning}, pruned LLMs show an obvious degradation in trustworthiness after pruning.
For example, Gemma-1.1-7B and LLaMA-3.1-8B exhibit noticeable declines, with LLaMA-3.1-8B dropping by nearly 10\% in robustness. 
These findings suggest that pruning substantially reduces model trustworthiness across multiple dimensions.

\noindent
\textbf{Semi-structured pruning increases degradation.}
As shown in Table~\ref{tab:pruning}, applying semi-structured pruning further reduces trustworthiness compared to unstructured pruning. In particular, the performance under 2:4 sparsity is worse than under 4:8 sparsity, indicating that more restrictive sparsity patterns introduce greater degradation.
This suggests that rigid pruning constraints can distort important representation subspaces, thereby amplifying the loss of trustworthiness across multiple dimensions.
\textit{Considering that unstructured pruning offers little inference speedup and semi-structured pruning leads to low trustworthiness, we do not recommend using pruned models as trustworthy SLMs}.

\begin{table}[!t]
    \centering
    \resizebox{\linewidth}{!}{
\begin{tabular}{l|ccccc}
\toprule
\textbf{Model}          & \textbf{Ethics} & \textbf{Privacy} & \textbf{Robustness} & \textbf{Fairness} & \textbf{Overall} \\ \midrule
\textbf{Qwen2.5-1.5B}   & 77.52\%         & 40.89\%          & 69.38\%             & 71.82\%           & 64.90\%          \\ \midrule
\textbf{1.5B-INT8-GPTQ} & 77.68\%         & 41.08\%          & 69.13\%             & 71.85\%           & 64.93\%          \\
\textbf{1.5B-INT4-GPTQ} & 78.63\%         & 40.38\%          & 67.34\%             & 69.07\%           & 63.86\%          \\
\textbf{1.5B-INT3-GPTQ} & 77.31\%         & 40.72\%          & 67.01\%             & 68.23\%           & 63.32\%          \\ \midrule
% \midrule
\textbf{Qwen2.5-3B}     & 80.13\%         & 40.48\%          & 68.79\%             & 29.99\%           & 54.85\%          \\ \midrule
\textbf{3B-INT8-GPTQ}   & 80.26\%         & 40.35\%          & 68.70\%             & 29.54\%           & 54.71\%          \\
\textbf{3B-INT4-GPTQ}   & 80.43\%         & 44.98\%          & 68.75\%             & 29.82\%           & 56.00\%          \\
\textbf{3B-INT3-GPTQ}   & 80.02\%         & 41.32\%          & 68.65\%             & 29.13\%           & 54.78\%          \\ \bottomrule
\end{tabular}

    }
    \caption{Trustworthiness assessment under different quantized bits for Qwen2.5 models.}
    \label{tab:quant_bits}
    \vspace{-0.3cm}
\end{table}

\begin{table*}[!t]
    \centering
    \resizebox{0.6\linewidth}{!}{
\begin{tabular}{l|ccccc}
\toprule
\textbf{Model}                           & \textbf{Ethics} & \textbf{Privacy} & \textbf{Robustness} & \textbf{Fairness} & \textbf{Overall} \\ \midrule
\textbf{h2o-danube3-500m-Chat}           & 61.92\%         & 29.02\%          & 56.85\%             & 43.27\%           & 47.76\%          \\
\textbf{MobiLlama-500m-Chat}              & 55.79\%         & 32.33\%          & 48.02\%             & 45.69\%           & 45.46\%          \\
\textbf{SmolLM2-360M-Instruct}           & 51.47\%         & 48.59\%          & 60.16\%             & 47.45\%           & 51.92\%          \\
\textbf{Qwen2.5-0.5B-Instruct}           & 65.71\%         & 27.56\%          & 62.42\%             & 48.99\%           & 51.17\%          \\ \midrule
\textbf{Qwen2.5-1.5B-Instruct}           & 77.52\%         & 40.89\%          & 69.38\%             & 71.82\%           & 64.90\%          \\
\textbf{Qwen2.5-1.5B-INT4-AWQ}       & 73.83\%         & 38.97\%          & 68.64\%             & 71.89\%           & 63.33\%          \\
\textbf{Qwen2.5-1.5B-INT4-GPTQ} & 78.63\%         & 40.38\%          & 67.34\%             & 69.07\%           & 63.86\%          \\
\textbf{Qwen2.5-1.5B-INT8-GPTQ} & 77.68\%         & 41.08\%          & 69.13\%             & 71.85\%           & 64.93\%          \\ \bottomrule
\end{tabular}
    }
    \caption{Trustworthiness comparison among pre-trained SLMs and compressed LLMs.}
    \label{tab:comparison}
    \vspace{-0.3cm}
\end{table*}

\begin{table*}[!t]
    \centering
    \resizebox{0.55\linewidth}{!}{
\begin{tabular}{l|ccccc}
\toprule
\textbf{Model}      & \textbf{Ethics} & \textbf{Privacy} & \textbf{Robustness} & \textbf{Fairness} & \textbf{Overall} \\ \midrule
\textbf{Qwen2.5-7B} & 83.78\%         & 41.53\%          & 67.05\%             & 28.33\%           & 55.17\%          \\
\textbf{Qwen2.5-3B} & 80.13\%         & 40.48\%          & 68.79\%             & 29.99\%           & 54.85\%          \\
\textbf{KD-3B} & 82.12\%         & 41.36\%          & 69.88\%             & 32.45\%           & 56.45\%          \\
% \textbf{KD-Math-3B} & 82.24\%         & 38.92\%          & 68.83\%             & 33.73\%           & 55.93\%          \\ 
\bottomrule
\end{tabular}
    }
    \caption{Exploration of distillation impact.}
    \label{tab:distillation}
    \vspace{-0.3cm}
\end{table*}

\subsection{Quantization Influence}

\noindent
\textbf{Quantization has relatively minor influence, especially for larger models.}
As shown in Table~\ref{tab:quantization}, quantization maintains the trustworthiness of Qwen2.5-7B with minimal degradation. We further extend our evaluation to 4-bit quantization on smaller Qwen models and the LLaMA-3.2 series.
According to Table~\ref{tab:quantization}, the quantized models generally exhibit comparable performance to their full-precision counterparts, especially in the case of larger models. 
For instance, the Qwen2.5-0.5B model experiences a moderate ~4\% drop under AWQ quantization, whereas GPTQ yields more stable results, even improving the accuracy. 
For models larger than 1.5B, the decrease in trustworthiness remains minimal (less than 2\%).
% indicating that larger models are inherently more robust to quantization.
\textit{These findings highlight quantization as a viable strategy for building trustworthy SLMs, with larger-scale models demonstrating greater resilience to low-bit compression.}

\noindent
\textbf{GPTQ offers more reliable trustworthiness than AWQ.}
As presented in Table~\ref{tab:quantization}, models quantized using GPTQ typically achieve higher accuracy in trustworthiness evaluations compared to those quantized with AWQ.
This trend holds across most of the settings in our experiments. 
Interestingly, we observe that for certain models, such as LLaMA-3.2-3B and Qwen2.5-3B, GPTQ quantized variants even outperform their FP16 counterparts. 
These results further demonstrate that quantized models, particularly when using GPTQ, can not only reduce memory consumption but also maintain or even enhance trustworthiness, highlighting the potential of low-bit quantization in building efficient and reliable SLMs.

\noindent
\textbf{Quantization showcases robustness towards different compression ratios.}
We further analyze the effect of different quantization levels on trustworthiness.
As shown in Table~\ref{tab:quant_bits}, GPTQ exhibits stable performance across 3–8 bits on Qwen2.5, with no systematic degradation as the bit-width decreases.
This trend suggests that low-bit quantization largely preserves the reliability-related behaviors learned by the full-precision model, and the trustworthiness scores are relatively insensitive to the compression ratio within this range.
Overall, these results support our conclusion that quantization (especially GPTQ) is a robust and practical strategy for obtaining trustworthy SLMs under varying efficiency constraints.

\subsection{Directly Using SLMs or Compressing Larger LLMs?}
Based on our analysis of pruning and quantization effects on LLMs, we focus on quantization as the primary technique to compare compressed LLMs against directly pre-trained SLMs.
In this subsection, we evaluate several pre-trained SLMs and quantized versions of larger models. Specifically, we examine four pre-trained SLMs with fewer than 1B parameters to provide a comprehensive perspective.

\noindent
\textbf{Performance Analysis.}
As shown in Table~\ref{tab:comparison}, these pre-trained SLMs generally achieve lower trustworthiness scores, typically around 50%.
In contrast, we apply 4-bit quantization to Qwen2.5-1.5B, which can reduce weight memory by approximately 4× and potentially deliver up to 3× speedup with optimized kernels~\cite{lin2023awq}.
This makes the quantized 1.5B model comparable in practical deployment efficiency to smaller SLMs (e.g., 0.5B models), while retaining stronger representations.
Consequently, the quantized Qwen2.5-1.5B model achieves substantially higher trustworthiness (around 63\%) than the pre-trained SLMs.
This gain is driven by the minimal degradation introduced by quantization and the stronger base capability of the 1.5B model compared to smaller counterparts (e.g., Qwen2.5-0.5B).

\noindent
\textbf{Memory and Latency Discussion.}
On RTX-4090 with batch=1 decoding, generation latency is often memory-bound. Although the 1.5B model has \textasciitilde3× more parameters, INT4 quantization reduces weight memory traffic by \textasciitilde4×, which can largely offset the increased parameter count in practice. As a result, Qwen2.5-1.5B-INT4-GPTQ typically exhibits comparable ms/token to Qwen2.5-0.5B FP16 (often within \textasciitilde1.2–1.5×), rather than being strictly slower, while achieving markedly better trustworthiness.
Moreover, quantization offers a flexible knob to produce SLMs of different sizes and efficiency levels by adjusting bit-width, enabling smooth trade-offs between memory, speed, and trustworthiness.

From this analysis, we derive an important insight: instead of relying solely on small models trained from scratch, \textbf{leveraging quantization on larger pre-trained LLMs can yield compact, efficient, and more trustworthy SLMs}.

\subsection{Distillation Enhancement}
We explore whether knowledge distillation can improve SLM trustworthiness. Specifically, Qwen2.5-3B is distilled from Qwen2.5-7B using the Alpaca dataset, with results listed in Table~\ref{tab:distillation}. 
The distilled model shows improvements across all four trustworthiness categories, demonstrating the benefit of transferring knowledge from a more reliable model.
We expect that distilling from larger and more trustworthy teacher models could yield even greater improvements in the reliability of small language models.
\section{Conclusion}

This work explores effective strategies for developing trustworthy Small Language Models (SLMs) through a systematic empirical study. Our findings reveal three important observations:
(1) Quantization offers a more reliable means than pruning for preserving model trustworthiness;
(2) Compressing a well-aligned large model via quantization results in more robust and adaptable SLMs than directly using pre-trained small models;
(3) Knowledge distillation serves as a complementary approach to further improve the reliability of compact models.
Overall, our study sheds light on practical and scalable pathways to build efficient SLMs without compromising their trustworthiness, and lays the foundation for future efforts in this direction.

% \vfill\pagebreak
% \clearpage

\bibliographystyle{IEEEbib}
\bibliography{refs}

\end{document}